\documentclass[conference]{IEEEtran}

\IEEEoverridecommandlockouts
\usepackage{subfigure}

\usepackage{cite}
\usepackage{amsmath,amssymb,amsfonts}
\usepackage{algorithmic}
\usepackage{graphicx}
\usepackage{textcomp}
\usepackage{xcolor}
\def\BibTeX{{\rm B\kern-.05em{\sc i\kern-.025em b}\kern-.08em
    T\kern-.1667em\lower.7ex\hbox{E}\kern-.125emX}}

\begin{document}

\title{CloudVision: DNN-based Visual Localization of Autonomous Robots using Prebuilt LiDAR Point Cloud}

\author{\IEEEauthorblockN{Evgeny Yudin,
Pavel Karpyshev,
Mikhail Kurenkov, 
Alena Savinykh,
 \\
Andrei Potapov, 
Evgeny Kruzhkov, and 
Dzmitry Tsetserukou}
\IEEEauthorblockA{\textit{ISR Laboratory, Skolkovo Institute of Science and Technology, Moscow, Russia}}
\IEEEauthorblockA{$\{$Evgeny.Yudin, Pavel.Karpyshev, Mikhail.Kurenkov, Alena.Savinykh, \\
Andrei.Potapov, Evgeny.Kruzhkov, D.Tsetserukou$\}$@skoltech.ru}}

\maketitle

\begin{abstract}

In this study, we propose a novel visual localization approach to accurately estimate six degrees of freedom (6-DoF) poses of the robot within the 3D LiDAR map based on visual data from an RGB camera. The 3D map is obtained utilizing an advanced LiDAR-based simultaneous localization and mapping (SLAM) algorithm capable of collecting a precise sparse map. The features extracted from the camera images are compared with the points of the 3D map, and then the geometric optimization problem is being solved to achieve precise visual localization. Our approach allows employing a scout robot equipped with an expensive LiDAR only once --- for mapping of the environment, and multiple operational robots with only RGB cameras onboard --- for performing mission tasks, with the localization accuracy higher than common camera-based solutions. The proposed method was tested on the custom dataset collected in the Skolkovo Institute of Science and Technology (Skoltech). During the process of assessing the localization accuracy, we managed to achieve centimeter-level accuracy; the median translation error was as low as 1.3 cm. The precise positioning achieved with only cameras makes possible the usage of autonomous mobile robots to solve the most complex tasks that require high localization accuracy.

\end{abstract}


\section{Introduction}

\subsection{Motivation}

Today, mobile robotics is one of the fastest growing areas of research. At the moment, both the scientific community and the industry are interested in obtaining modern and reliable solutions in this area. Regarding industry, the mobile robots market is valued at 1.61 billion dollars in 2021 and is expected to grow to 22.15 billion dollars by 2030 \cite{tmirob}. The interest of the academic community in the field of robotics is proven by a massive increase in the number of papers, journals, and conferences devoted to all areas and applications of autonomous machines. The efforts are aimed at ensuring accurate, efficient and safe operation of autonomous robots by providing them with reliable algorithms.

\begin{figure} [t] 
\begin{center}
\includegraphics[width=8.4 cm]{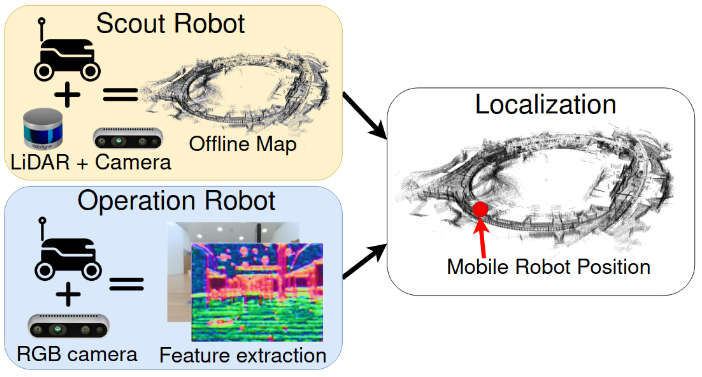}
\caption{Concept of visual localization pipeline within LiDAR map using two types of robots: a scout robot for mapping and an operation robot for localization.}
\vspace{-1.5em}
\label{scheme_concept}
\end{center}
\end{figure}

\subsection{Problem Statement}

\begin{figure*}[!t] 
\begin{center}
\includegraphics[width=01\textwidth]{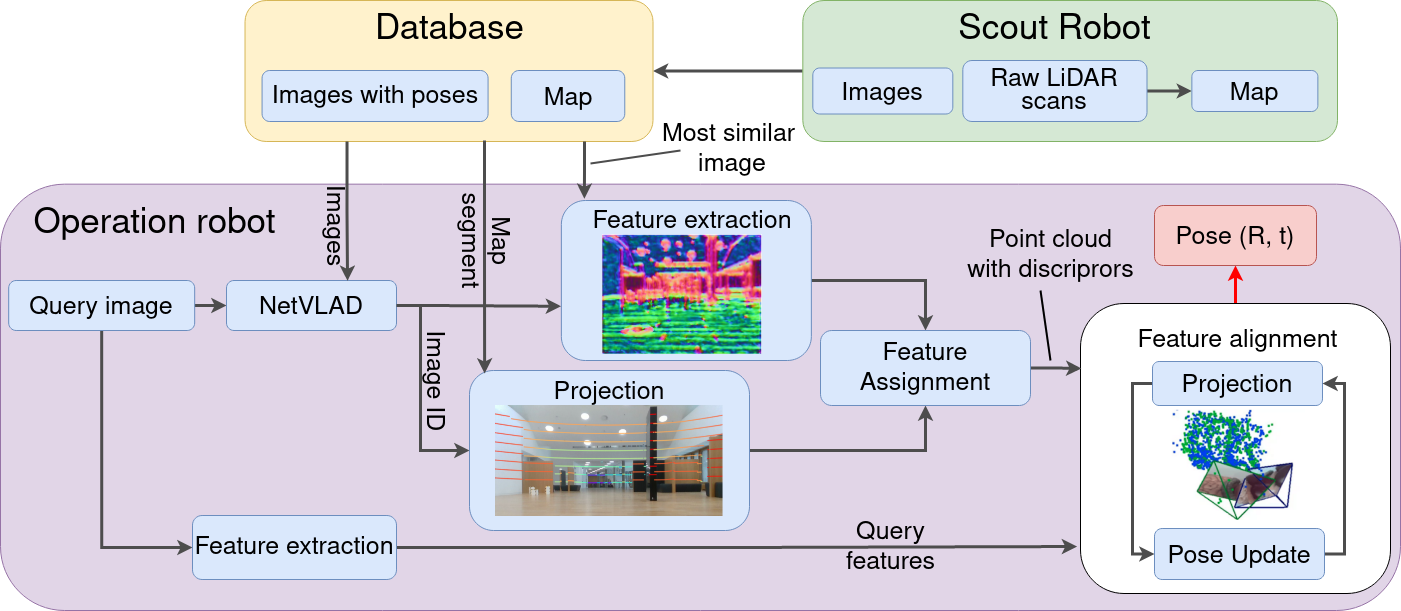}
\caption{Proposed visual localization pipeline in 3D LiDAR map.}
\vspace{-1.5em}
\label{scheme_all}
\end{center}
\end{figure*}

An essential task that arises during the operation of a mobile robot is localization. It is important for a robot to determine its position at any given moment in order to plan the path, avoid collisions with dynamic and static obstacles, etc. The common way to solve this problem is to use algorithms that process data from cameras, LiDARs, or a combination of both.

Cameras are the most widespread perception sensors due to their cheapness and ability to obtain dense information on the environment, including colors, depth, and shapes. However, visual localization algorithms show less accurate results than LiDAR-based solutions. To avoid the accuracy decrease of such methods, the input image sequence must be visually heterogeneous and rich in terms of features; unfortunately, such requirements are far from the real-world conditions. Nevertheless, cameras have found an application in many robotic projects due to their low cost. Thus, camera-based technologies are suitable for solving crucial perception operation tasks. On the other hand, they suffer from relatively low accuracy and robustness \cite{toft2020long}. 

The main advantage of LiDARs, in turn, is obtaining highly reliable data on the distance to objects without preprocessing, regardless of changes in lighting and changing seasons. The most accurate solutions for localization are LiDAR-based approaches \cite{elhousni2020survey}. However, LiDAR data is vulnerable to change of weather conditions (snow, rain, mist). Moreover, LiDARs are an expensive solution that is difficult to scale due to the high price for such sensors.

For these reasons, currently there is no single approach to designing a sensor setup for the localization problem. On the one hand, LiDARs provide the highest accuracy, but are costly. On the other hand, cameras are affordable, but do not provide high accuracy.

\subsection{Related Works}

\textbf{LiDAR-based simultaneous localization and mapping (SLAM)} algorithms are able to achieve high localization accuracy and generate a high-quality sparse 3D map of the environment. According to the results on KITTI benchmark \cite{geiger2012we}, the most accurate SLAM algorithms are based on Lidar Odometry and Mapping (LOAM) method  \cite{zhang2014loam}, utilizing LiDAR as the main source of information. The core idea of this algorithm is the extraction of edges for geometrical surfaces depicted on the LiDAR scan.
When a new frame of the LiDAR scan is received, the algorithm computes the edges on it and matches them with the detected edges on the previous frame, thus obtaining the 3D map and estimating the robot trajectory and position. 
The main disadvantage of the LOAM method is the lack of loop closure detection, which leads to an increase in the drift trajectory error over time. 

Multiple papers are devoted to extending the LOAM approach to achieve better performance under different conditions. For example, LeGO-LOAM \cite{shan2018lego} improves the original approach using ground partitioning, point cloud segmentation, and advanced computational optimization. Segmentation allows detecting noisy points that may represent unreliable features and filter them out. Ground partitioning divides features into planar and edge, which makes the algorithm more reliable.
Another work based on LOAM, called F-LOAM \cite{wang2021f}, improves the computational efficiency by combining scan-to-scan match and scan-to-map refinement. 

Other approaches to LiDAR-based SLAM \cite{shi2020spsequencenet, yin2020cae} are based on Convolutional Neural Networks (CNNs) for point cloud processing. Although deep learning approaches show good results on open datasets in terms of odometry accuracy, they are unreliable in cases of drastic scene change, such as moving from indoor to outdoor scene. Thus, for robust operation, these approaches should be trained again on unseen scenes, which significantly reduces their applicability in real-world scenarios.

\textbf{Sensor Fusion Methods}, particularly camera and LiDAR data fusion, has found wide application for solving both localization and SLAM problems. Researchers \cite{zhang2015visual, chen2021lidar, graeter2018limo, seo2019tight} propose highly accurate localization approaches based on the LiDAR-camera data fusion. However, all of them require the presence of both camera and LiDAR in the sensor setup of each robot.

Yu et al. \cite{yu2020monocular} suggest applying a LiDAR map to estimate the camera position using visual localization based on matching 2D lines on the image and 3D lines in the point cloud map. This method has shown accurate localization results on small maps; however, it is inapplicable to larger areas. Several papers \cite{caselitz2016monocular, ding2017fusing} propose to compare 3D point clouds reconstructed from monocular camera images with LiDAR maps. This allows to solve the problem of estimating the real-world scale factor for trajectories and maps obtained with a monocular camera. However, the direct comparison of 3D LiDAR maps and reconstructions obtained by cameras is inconsistent due to their sparse structure. This, in turn, leads to a rough estimate of the camera pose.

Feng et al. \cite{feng20192d3d}  implemented a deep learning approach to compute the descriptors that allow direct matching of keypoints across an image and a 3D map. The proposed neural network is trained to match features on the specific 3D scene. Thus, this method is highly dependent on the training data and unable to achieve the reliable results on scenes not included in the training dataset.

\textbf{Visual Localization} methods can be divided into two subgroups based on their operation principle: image-based and structure-based. Image-based localization approaches rely solely on images, and do not require the storage of a 3D scene for localization in an explicit form. In particular, PoseNet \cite{kendall2015posenet} leverages a CNN, such as VGGNet or ResNet, to regress both camera position and orientation. Another approach, VLocNet \cite{valada2018deep}, extends the global pose regression neural network with an additional subnetwork to estimate the translation from the last frame, thereby additionally solving the problem of visual odometry. It is further extended by VLocNet++ \cite{radwan2018vlocnet++}, that is able to simultaneously estimate visual odometry, global localization, and perform semantic segmentation. The main disadvantage of these approaches is their heavily dependency on training data, which leads to scaling problems.

Image retrieval systems aim to find similar images to a query image (i.e., the current image obtained from the camera) among a dataset. State-of-the-art approaches aimed at solving the image retrieval problem mainly use trained global descriptors. For example, NetVLAD \cite{arandjelovic2016netvlad} is able to determine the most similar image to a query image with a high degree of invariance to changes in conditions (scale, illumination, etc.). Patch-NetVLAD approach \cite{hausler2021patch} solves the problem of localization by combining global and local descriptors, which allows calculating the position in a large area. Both these approaches have shown high operating speeds, but low accuracy for pose estimation, however, they excel at image retrieval tasks.

\begin{figure} [!t] 
\begin{center}
\includegraphics[width=7.4 cm]{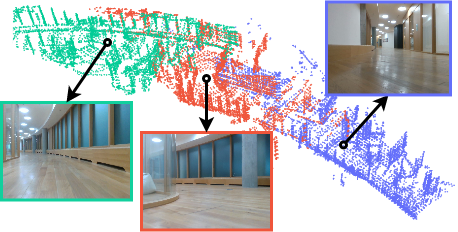}
\caption{Indexed point cloud according to database images.}
\vspace{-1.5em}
\label{cov}
\end{center}
\end{figure}

Structure-based methods calculate the pose of the camera in a reconstructed 3D map of the area. Most of them involve the Structure from Motion (SfM) \cite{schonberger2016structure, schonberger2016pixelwise} which is able to estimate the 3D structure of a scene from a set of 2D images. For localization, keypoints are extracted from the query image and matched with the 3D model using its descriptors. When the matches between the image and the 3D model are determined, the Perspective-n-Point problem is solved to calculate the position of the camera. This approach is able to provide highly accurate localization given an accurate 3D model, and a sufficient number of detected keypoints; however, with the growth of the 3D model, the computing efficiency significantly decreases. The large-scale scene problem can be solved by selecting a segment of the global 3D map, in which the query image is located. This can be done using GPS or image retrieval as a source of approximate location information.

The most outstanding modern approach to solving visual localization problem is PixLoc \cite{sarlin2021back}. It is a scene-invariant neural network that allows to extract dense features from images. PixLoc learns to distinguish pieces of images, that are most suitable for localization, through end-to-end learning from pixels to pose, and demonstrates exceptional generalization to new scenes by separating model parameters and scene geometry. 

Despite the fact that this approach shows most promising results compared to its alternatives, accuracy may still not be enough in cases that require precise robot localization. We propose to create a visual localization pipeline based on LiDAR maps, investigating the impact of replacing Structure from Motion to a map obtained by the state-of-the-art LiDAR SLAM algorithm. In our hypothesis, the precise LiDAR point clouds would improve the localization accuracy and overall robustness of the pipeline.

\subsection{Contribution}
We propose and evaluate a novel approach to solving the problem of visual localization by matching camera data with a prebuilt 3D LiDAR map. The concept for this approach is to use two types of robots. The first one collects a 3D map and database images by means of camera and LiDAR, and the robots of the second type are equipped with only a camera and leverages the prebuilt LiDAR map for visual localization. The layout of the proposed concept is shown in Fig. \ref{scheme_concept}.

In the scope of this research, we analyze the existing LiDAR-based mapping approaches and choose the optimal one for collecting a 3D map. We develop an algorithm that allows to collect a map with information about the visibility of each point based on the camera's field of view. Next, we adapt the 3D map for the Pixloc pipeline, and conduct a series of experiments to determine the accuracy of indoor localization using the proposed approach.

\begin{figure*}[t]
\centering
\subfigure[Raw LiDAR scan]{
\includegraphics[width=.31\textwidth]{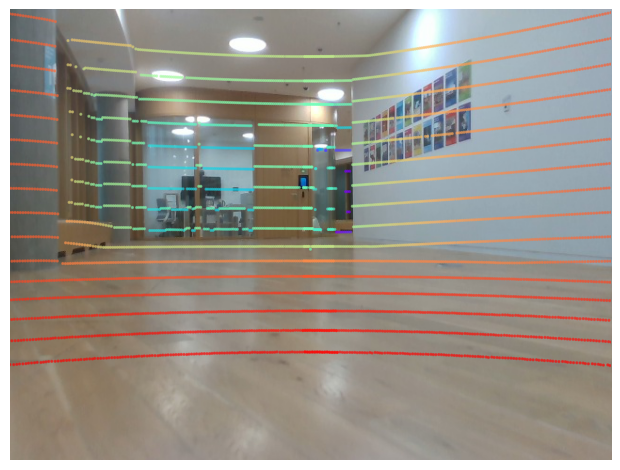}
}
\subfigure[Structure from Motion]{
\includegraphics[width=.31\textwidth]{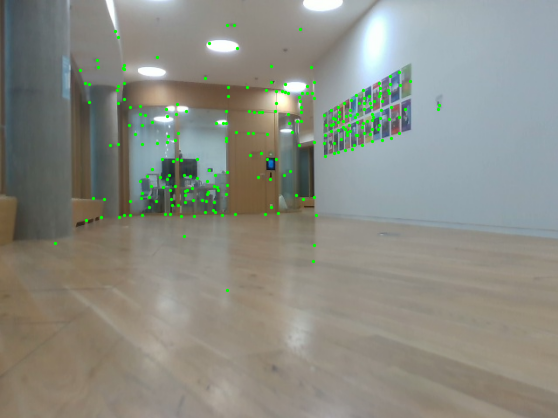}
}
\subfigure[LiDAR map]{
\includegraphics[width=.31\textwidth]{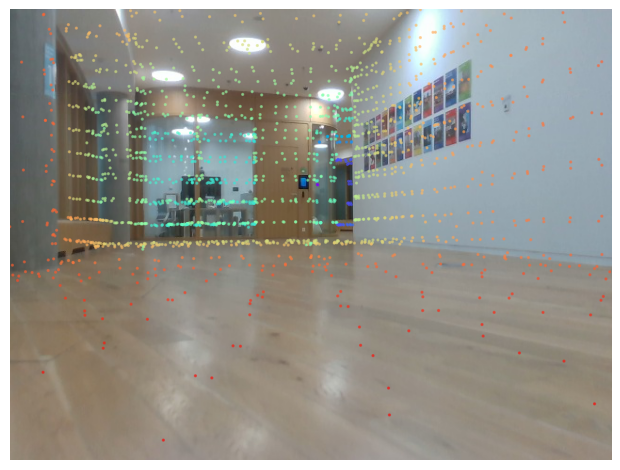}
}
\caption{Points projection of different 3D structures: (a) Raw LiDAR scan. (b) Structure from Motion. (c) LiDAR MAP.}
\label{fig:projection_all}
\end{figure*}

\section{Method of Visual Localization}

Our pipeline can be divided into three main steps. The first step is to create a 3D structure of the environment using an advanced LiDAR-based SLAM algorithm. During the second step, the 3D structure segment is selected using image retrieval, which allows reducing the search area. In the last step, we use the dense feature extractor and classical geometric optimization to get the 6-DoF robot pose $(\textbf{R}, \textbf{t})$, where $\textbf{R}$ is a rotation matrix and $\textbf{t}$ is a translation vector of the robot. The full pipeline is shown in Fig. \ref{scheme_all}. 

\subsection{Obtaining 3D Map}

Analyzing the results of LiDAR-based SLAM algorithms on the KITTI benchmark, it is noticeable that LOAM algorithms family occupies a leading position in terms of odometry accuracy. To build a 3D map, we employ an advanced implementation of LOAM, called A-LOAM \cite{hkust-aerial-robotics} since it is open source.
It provides precise localization without requiring high accuracy LiDAR ranging and inertial measurements. The high 3D map accuracy is achieved by reducing its update rate for proper matching and point cloud registration. In addition, the algorithm is able to work in real time, that makes it perfectly suitable for building a 3D map on a scout robot. In our approach, we extend A-LOAM with our algorithm for projection of co-visible map points.

To estimate the position of the camera, it is necessary to find the map segment corresponding to the current query image and project the map points onto the image. The problem is that it needs to project only those map segment points that are visible to the camera, excluding the rest, e.g., located around the corner. To solve this problem, we modify the stage of obtaining a map in the A-LOAM algorithm: we assign an index of the database image to each point of the LiDAR scan corresponding to that image. The result of point cloud indexing is shown in Fig. \ref{cov}. This approach allows to effectively filter the points that are not visible from the camera, knowing the common database images. The projection of map points on the image is shown in Fig. \ref{fig:projection_all}(c).

\subsection{Image Retrieval}

The computational efficiency of structure-based visual localization decreases with the growth of the 3D structure. A preliminary estimation of a robot pose can help to significantly speed up the calculation of the exact pose inside a small map segment. The approximate pose estimation is carried out either using GPS or image retrieval techniques. However, it is not always possible to obtain the GPS signal, especially in case of indoor operation. Therefore, the image retrieval method is more versatile, and was used in our work to estimate the location, at which the query image was captured, using the location of the most visually similar image from a database.

In our approach, to solve the image retrieval problem we use a CNN-based architecture called NetVLAD \cite{arandjelovic2016netvlad}, that is trained in an end-to-end manner specially for the place recognition task. This network extracts global descriptors from images, and the comparison of such descriptors provides an identification of the most similar images. Thus, having a database of images, it is possible to quickly evaluate the place where the query image was made using NetVLAD. The high speed of such evaluation is possible due to the small size of descriptors, and the ability to calculate them for the database images preliminarily. Thereby, we can extract the most similar image from the database to the current query image and, using the previously described algorithm, define the co-visible map points. The described approach is used for efficient and fast filtering of a 3D scene, which significantly limits the search area. In our work, we use NetVLAD weights pretrained on the Pittsburgh (Pitts30k) dataset \cite{torii2013visual}.

\subsection{Pose Estimation}

To estimate the robot pose, we utilize the algorithm proposed in Pixloc \cite{sarlin2021back}. Basically, this method estimates the pose where the query image was captured by aligning its features to the database image in the 3D structure.
Using NetVLAD global descriptors, we find the image from the database that is the most similar to the current query image. After that, we use the Pixloc convolutional neural network to extract dense features from the database image. The network is invariant to the scene, and is able to work equally well in both indoor and outdoor scenes without retraining. In our work, we use the Pixloc network pretrained with MegaDepth dataset \cite{li2018megadepth}.
Next, the points of the filtered 3D structure are projected onto the retrieved database image, and 2D descriptors obtained from Pixloc are assigned to each 3D point. Thus, we get a point cloud with image descriptors. The next step is minimizing the difference in appearance between the query image and database image. Pixloc's dense features are also extracted from the query image, and certain points with a descriptor are projected onto it. Thus, the position, from which the query image was captured, is estimated by minimizing the error using the Levenberg-Marquardt algorithm \cite{levenberg1944method, marquardt1963algorithm} by aligning Pixloc features from the query image and Pixloc features and corresponding 3D points from the database image.

\section{System Overview}

\subsection{Hardware and Software Architecture}

\begin{figure} [!t] 
\begin{center}
\includegraphics[width=7.4 cm]{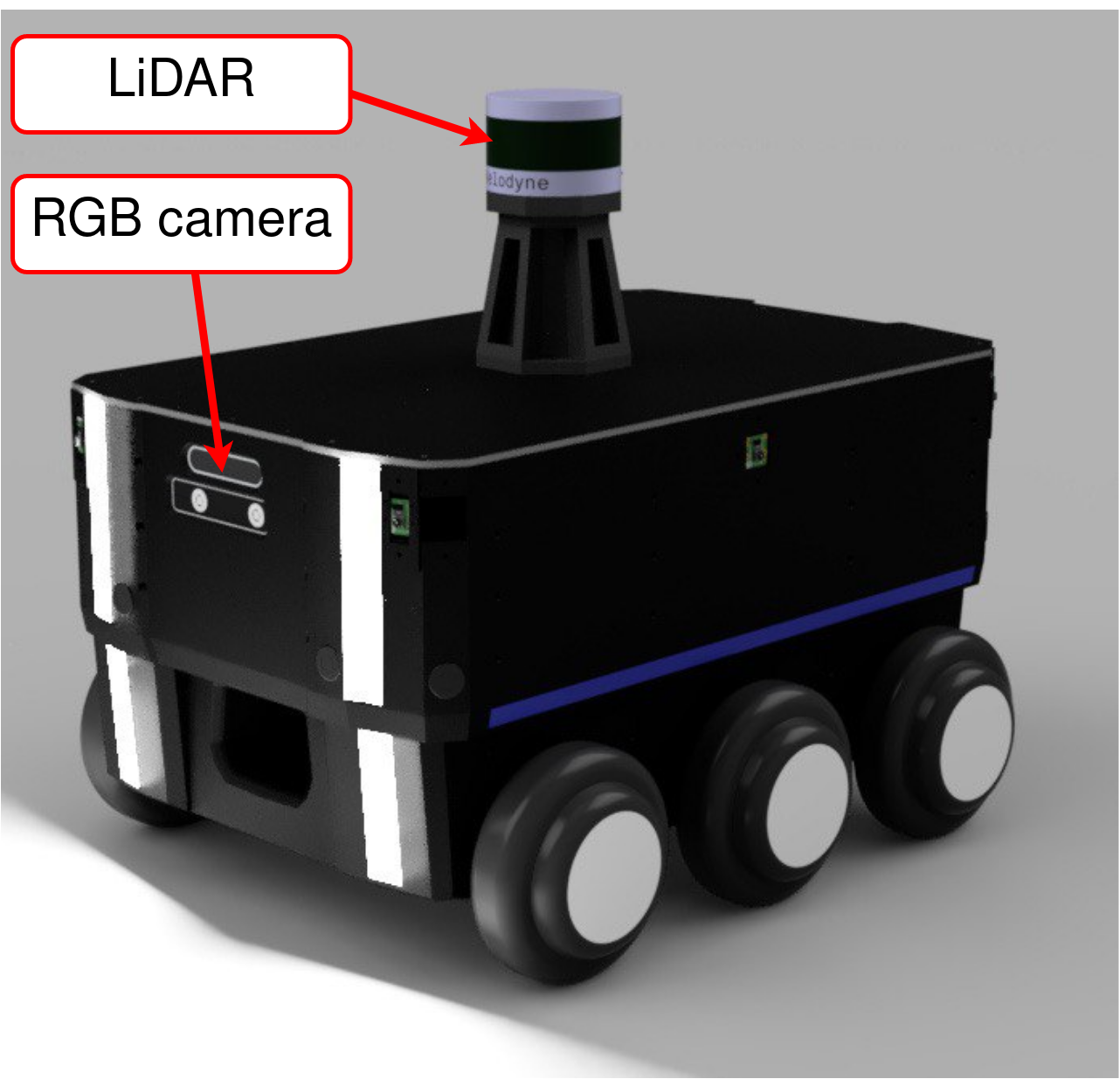}
\caption{HermesBot autonomous platform equipped with Velodyne VLP-16 LiDAR and Intel RealSense D435 RGB-D camera.}
\vspace{-1.5em}
\label{hermes_render}
\end{center}
\end{figure}

To collect the dataset, the HermesBot autonomous platform \cite{protasov2021cnn}, depicted in Fig. \ref{hermes_render}, was utilized. The robot is equipped with all sensors required for a comprehensive perception system. The sensor setup includes a Velodyne VLP-16 LiDAR, and two Intel RealSense D435 RGB-D cameras, mounted on the front and back sides of the robot. The intrinsic parameters of RealSense cameras were obtained prior to data collection. The rigid body transformation between the LiDAR and the camera was preliminarily determined by the geometry of the robot CAD model and refined using the calibration algorithm \cite{tsai2021optimising}.

The platform is controlled by a tandem of two computing units: an Intel NUC computer with an Intel Core i7 processor for general control and data processing, and an NVIDIA Jetson Xavier module for neural networks operation, e.g., Pixloc dense feature extraction. All sensors and modules are connected using the ROS framework.

\subsection{Dataset Collection}

To evaluate the proposed approach, we collected the indoor set of LiDAR and visual data in the campus of Skoltech. The dataset consists of two sequences collected on similar closed trajectories. The length of each trajectory was 380 meters, with start and finish at the same point. Sequence 1 corresponds to the activity of the first scout robot and contains both LiDAR scans and camera data, and sequence 2 corresponds to the operation robot and comprises visual information involved in localization algorithm testing, and LiDAR data for obtaining ground truth. Sequence 1 was used to build the 3D map and collect the image database, while sequence 2 was employed for estimating the localization accuracy using the proposed method.

\begin{figure} [tbp] 
\begin{center}
\includegraphics[width=7.4 cm]{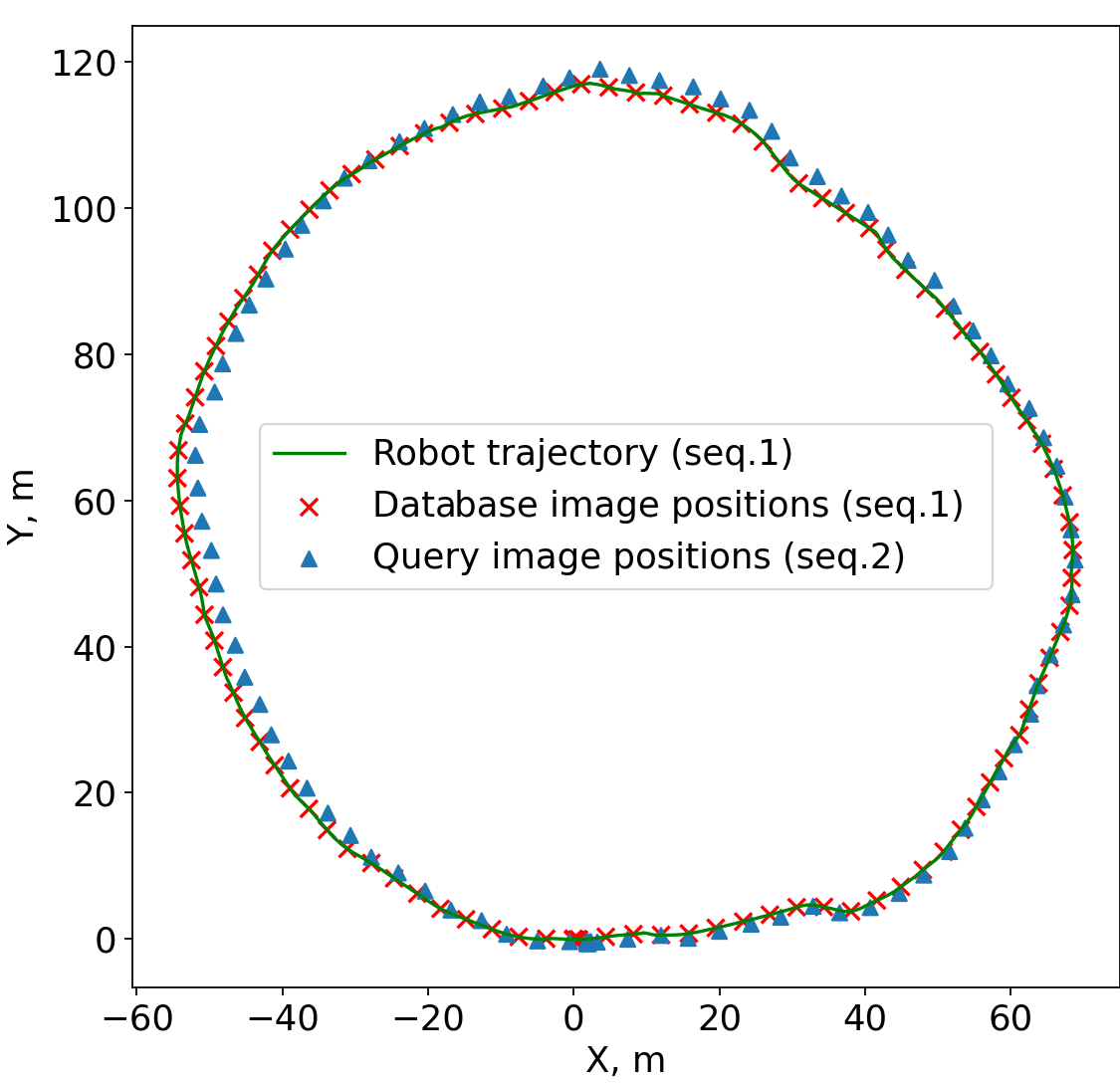}
\caption{ Bird's-eye view of robot trajectory and ground truth camera poses for scout and operation sequences.}
\vspace{-1.5em}
\label{traj}
\end{center}
\end{figure}

For sequence 1, 4966 LiDAR scans were recorded. Visual data was recorded at 30 frames per second and resolution of 640x480, in RGB format. Each sequence consists of approximately 15000 camera frames, from which we have selected 300 frames per sequence, uniformly distributed along the entire length of the sequence. These images from sequence 1 were added to the database for subsequent NetVLAD image retrieval, providing approximate pose estimation. The selected images from sequence 2 were used as query images for position estimation and accuracy evaluation. This number of images was chosen as a trade-off between system operating speed and preliminary localization accuracy. The use of 300 images allows for estimating the approximate robot position with 2-meter accuracy, which allows to fully evaluate the performance of the proposed approach. For obtaining robot trajectory and ground truth poses used advanced LiDAR-based SLAM algorithm. Each camera pose contains three linear and three angular coordinates for the future 6-DoF pose estimation. 
Bird's-eye view of robot trajectory and ground truth camera poses for scout and operation sequences are depicted in Fig. \ref{traj}.

\section{Experiments and Result Discussion}

In order to evaluate the proposed approach, two sets of experiments were carried out. The aim of the first experiment was to demonstrate that the scale of the 3D map, obtained using the LiDAR-based SLAM approach, corresponds to real-world dimensions through localization accuracy estimation. The second set of experiments was aimed at assessing the localization accuracy of the proposed algorithm and comparing it with state-of-the-art visual approaches.

\subsection{Mapping Evaluation}

In the scope of this research, A-LOAM is not only used to build a 3D map, but also to assign ground truth camera positions for database and query images. For these reasons, it is important to verify the accuracy of A-LOAM, since this algorithm does not have a loop closure module.

The lack of loop closure detection in SLAM leads to inability to minimize drifting error. Therefore, we need to calculate the relative translation error. Since the start and end positions of the recorded trajectory are almost in the same point, we were able to physically measure the distance between them with millimeter accuracy. The measurements were performed similarly to the ones used in the TUM dataset \cite{sturm2012benchmark}. We compared the real displacement of the robot in the final position to the shift obtained from the trajectory estimated by A-LOAM. The relative translation error in the experiment constituted 0.13\%, while the entire length of the sequence was 380 m. The absolute translation error was 0.49 m.

The proposed approach estimates the robot pose using the most similar database image, therefore the error of the SLAM algorithm will be taken into account only in the distance relative to the positions from which the images were taken. It means that this translation error allows evaluating the accuracy of localization with high precision, since it accumulates during the entire sequence. In order to prove the consistency and uniformity of error accumulation, we ran the LiDAR-based SLAM on short sequences, physically measuring the distance traveled by the robot. Given that the average distance between the query image and the closest one database image found from the image retrieval is 2 meters, we conducted 10 launches of the robot on a trajectory 3 meters long and determined the absolute error, which averaged at 0.5 cm. This error makes it possible to estimate the accuracy of localization with centimeter accuracy.

\begin{table*}[ht]
\caption{Performance of Localization}
\begin{center}
\centering
\begin{tabular}{|c|c|c|c|c|l}
\cline{1-5}
Method &
  \begin{tabular}[c]{@{}c@{}}Median translation\\  error, cm\end{tabular} &
  \begin{tabular}[c]{@{}c@{}}Median rotation\\  error, deg.\end{tabular} &
  \begin{tabular}[c]{@{}c@{}}Percentage of images\\  at (5 cm, 2 deg.)\end{tabular} &
  \begin{tabular}[c]{@{}c@{}}Number of points\\  in millions\end{tabular} &
   \\ \cline{1-5}
hloc + SfM                   & 4.5 & 0.29 & 55.3 & \textbf{0.03} &  \\ \cline{1-5}
Pixloc + SfM               & 3.1 & 0.21 & 64.7 & \textbf{0.03} &  \\ \cline{1-5}
Pixloc + Raw LiDAR data (ours)   & 1.8 & 0.13 & 97.3 & 90   &  \\ \cline{1-5}
Pixloc + LiDAR Map (ours) & \textbf{1.3} & \textbf{0.09}  & \textbf{99.3} & 0.72 &  \\ \cline{1-5}
\end{tabular}
\label{table:table1}
\end{center}
\end{table*}

\subsection{Localization Evaluation}

\begin{figure} [!t] 
\begin{center}
\includegraphics[width=8.4     cm]{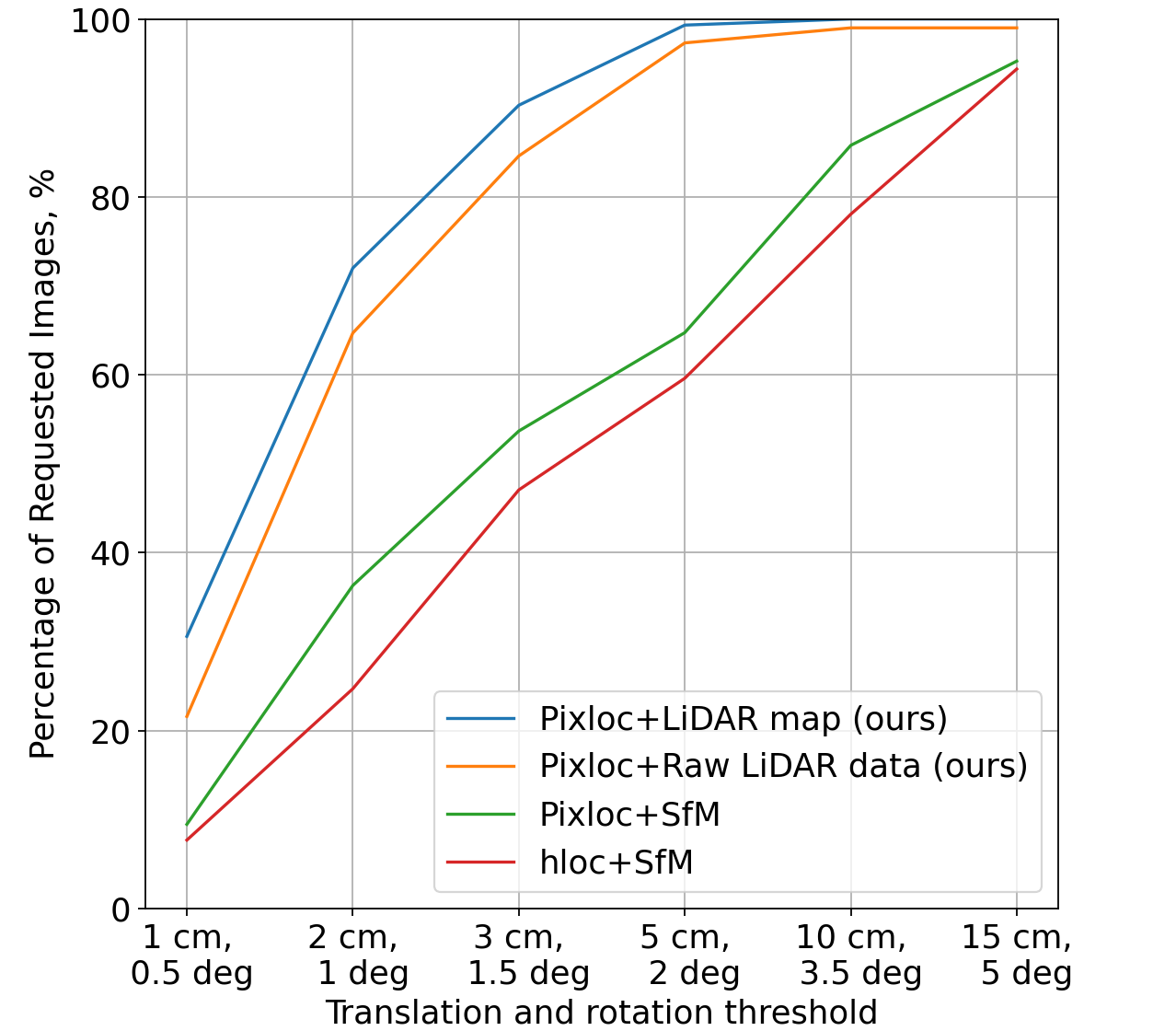}
\caption{Percentage of query images within certain limits for translation and rotation error.}
\vspace{-1.5em}
\label{plot1}
\end{center}
\end{figure}

To assess the localization accuracy, we estimate the median translation and rotation error \cite{kendall2015posenet} of the robot position in the sequence 2 for 300 query images. Fig. \ref{plot1} demonstrates the selected accuracy error thresholds and the percentage of poses estimated within this accuracy range. For the proposed approach validation, we compare its localization accuracy with the accuracy of state-of-the-art visual methods based on Structure from Motion, Pixloc and hloc \cite{sarlin2019coarse}, and with ablation modification of our method, in which the 3D map is replaced by simple raw LiDAR scans. The results of projecting 3D points on the image for different methods are shown in Fig. \ref{fig:projection_all}.

For Pixloc and hloc, the SfM models are reconstructed from 600 database images included in sequence 1. To build the model, we initially extract features from database images using Superpoint \cite{detone2018superpoint} and match them using SuperGlue \cite{sarlin2020superglue}. 
To assign the positions of the cameras in the model, we use the previously obtained ground truth poses obtained by A-LOAM. Then, we triangulate the extracted points with respect to these poses using their match information.
The resulting 3D structure contains 28000 points.

According to the experimental results shown in Fig. \ref{plot1}, the proposed approach is able to achieve centimeter accuracy; its median translation and rotation errors were 0.13 cm and 0.09 deg. respectively  according to the results in Table \ref{table:table1}. Moreover, the proposed method outperforms the state-of-the-art visual localization approaches in terms of accuracy, at least 2.5 times. 
The authors consider that such superiority of the proposed method is explained by the usage of much denser maps obtained by LiDAR SLAM. SfM usually provides sparse point clouds due to unstable keypoints extraction and matching, which is clearly much more scene-dependent than LiDAR technology (Fig. \ref{fig:projection_all}(b)). 

The superiority over the raw LiDAR approach is due to the fact that the scans are projected onto the database image in lines (Fig. \ref{fig:projection_all}(a), and that negatively affects the geometric optimization of the pose. Although the raw data contains the accurate description of the environment, the proposed approach provides greater accuracy, and a significantly smaller number of points in the structure, due to a more uniform distribution of points in the 3D map.

\section{Conclusions}

We have developed a novel localization approach for visual localization of autonomous robot based on prebuilt  LiDAR point cloud map. This was achieved by adapting the existing visual localization algorithm called Pixloc to use explicit point clouds instead of Structure from Motion (SfM). The mapping pipeline was modified to using the A-LOAM LiDAR-based SLAM method with algorithm for choosing of co-visible 3D map points, and the localization part of Pixloc was adapted to accept LiDAR data instead of SfM.

To evaluate the proposed approach, we collected a dataset consisting of two trajectories in the same location, including both LiDAR and camera data. Sequence 1 was used for mapping, and the same algorithm was applied to the second trajectory and used as ground truth. After that, the proposed localization pipeline was applied to the second data sequence and compared with Pixloc and hloc state-of-the-art visual localization approaches. The experimental results have shown that the proposed pipeline significantly outperforms the existing camera-based methods in terms of accuracy. The median translation error was equal to 1.3 cm, which is approximately 2.5 times better than the best scoring Pixloc, that achieved a median translation error of 3.1 cm. The resulting pipeline is both accurate and scene-invariant due to the use of a neural network-based dense feature extractor.

The use of the proposed approach is able to significantly decrease the cost of a robot fleet compared with the one equipped with only LiDARs, and increase the accuracy and robustness compared to a camera-only fleet.

\section{Discussion and Future Work}

In the future, we plan to explore how changing the layout of objects in the area affects the localization quality, e.g., when the map differs from the environment perceived by the robot. Although Pixloc partially solves this problem by focusing on reliable features, it is necessary to ensure that the proposed approach is resistant to slight changes in the environment. We also plan to develop the possibility of updating the map, stitching different parts of it, detecting dynamic objects and not taking them into account in the model. 

Moreover, we plan to perform a more extensive evaluation of the proposed approach. This includes testing on additional datasets \cite{pavlov2019icevisionset}, in real-world conditions on various setups, for example, outdoor ground robots \cite{protasov2021cnn}, and UAVs \cite{kalinov2021impedance}, as well as examine the applicability of the approach to more sophisticated systems, e.g., modular two-wheeled rovers \cite{petrovsky2022two} and plant inspection robots \cite{karpyshev2021autonomous}.

\section*{Acknowledgements} 
The reported study was funded by CNRS and RFBR according to the research project No. 21-58-15006.


\bibliographystyle{IEEEtran} 
\bibliography{ref}
\end{document}